\documentclass[runningheads]{llncs}
\usepackage{graphicx}
\usepackage{lineno,hyperref}
\usepackage{amsmath}
\begin{document}
\title{Explaining with Attribute-based and Relational Near Misses: An Interpretable Approach to Distinguishing Facial Expressions of Pain and Disgust\thanks{The work presented in this paper was funded partially by grant FKZ 01IS18056~B, BMBF ML-3 (TraMeExCo) and partially by grant DFG (German Research Foundation) 405630557 (PainFaceReader). We would like to thank Mark Gromowski who helped us with preparing the used data set.}
}
\titlerunning{Explaining with Attribute-based and Relational Near Misses}
\author{Bettina Finzel\orcidID{0000-0002-9415-6254} \and
Simon P. Kuhn \and\\
David E. Tafler \and
Ute Schmid\orcidID{0000-0002-1301-0326}}
\authorrunning{B. Finzel et al.}
\institute{Cognitive Systems, University of Bamberg, An der Weberei 5, 96047 Bamberg, Germany\\
\email{\{bettina.finzel,ute.schmid\}@uni-bamberg.de, \{simon-peter.kuhn,david-elias.tafler\}@stud.uni-bamberg.de}}

\maketitle              % typeset the header of the contribution
\begin{abstract}
Explaining concepts by contrasting examples is an efficient and convenient way of giving insights into the reasons behind a classification decision. This is of particular interest in decision-critical domains, such as medical diagnostics. One particular challenging use case is to distinguish facial expressions of pain and other states, such as disgust, due to high similarity of manifestation. In this paper, we present an approach for generating contrastive explanations to explain facial expressions of pain and disgust shown in video sequences. We implement and compare two approaches for contrastive explanation generation. The first approach explains a specific pain instance in contrast to the most similar disgust instance(s) based on the occurrence of facial expressions (attributes). The second approach takes into account which temporal relations hold between intervals of facial expressions within a sequence (relations). The input to our explanation generation approach is the output of an interpretable rule-based classifier for pain and disgust. We utilize two different similarity metrics to determine near misses and far misses as contrasting instances. Our results show that near miss explanations are shorter than far miss explanations, independent from the applied similarity metric. The outcome of our evaluation indicates that pain and disgust can be distinguished with the help of temporal relations. We currently plan experiments to evaluate how the explanations help in teaching concepts and how they could be enhanced by further modalities and interaction.

\end{abstract}
\section{Introduction}

Reliable pain analysis is an open challenge in clinical diagnosis \cite{hassan2019automatic}. Pain is a highly individual experience influenced by multiple factors, e.g., the state of health \cite{fillingim2017individual}. If not expressed verbally, pain is often displayed through facial expressions and can be recognized by analyzing those \cite{siebers2016characterizing,chen2018automated,kunz_facial_2019}. While individuals who are fully conscious are usually able to communicate their level of pain to the medical staff, patients with limited communication skills, e.g., sedated patients in intensive care units or patients with neurodegenerative disorders such as dementia are often not able to express themselves properly \cite{achterberg2021chronic,rieger2019make}. Furthermore, pain can be easily confused with emotions such as disgust due to the similarity in facial expressions \cite{kunz2013pain}, which makes pain assessment and treatment a complex task. The quality of pain assessment highly depends on the health care management. Medical staff is often not experienced enough or trained sufficiently to recognize pain in individual patients \cite{tsai2018pain}. Pain assessment methods supported by technology that analyze facial expressions are therefore important tools to tackle this challenge in the future \cite{lautenbacher2022automatic}.

One possibility to help medical staff in their decision making processes is to provide support based on artificial intelligence (AI) systems. Especially in image-based medical diagnostics, machine learning algorithms can be applied to segment images or to propose a classification \cite{erickson2017machine}. However, for reasons of accountability and traceability of a diagnostic outcome, experts have to justify their decision based on the observations they made. This applies as well to machine learning models that are used as decision-support for critical tasks, such as choosing the right dose of medication to treat pain \cite{achterberg2021chronic}. However, machine learned decision models are often black boxes. That is, it is not transparent on what information the system's decision has been based on. With respect to transparency of algorithmic decisions, explainable artificial intelligence (XAI) is an important research field which aims at (1) making obscure and complex machine learning models transparent, or (2) using interpretable and inherently comprehensible machine learning approaches \cite{holzinger2017we,rudin2019stop,arrieta_explainable_2019,schwalbe2021taxonomy}. In both cases, the human expert gets an explanation that gives insights into why a certain decision was made by the learned model.

To make explanations efficient, they have to be tailored to the specific needs of stakeholders and situational demands \cite{miller_explanation_2017}. Mostly, explanations are given to justify why a certain example has been classified in a specific way. Such, so called \textit{local} explanations, are especially helpful to confirm or broaden an expert's opinion. In contrast, it might be helpful for domain experts to gain insight into the complex decision space underlying a medical concept, that is, get a more \textit{global} explanation for the patterns observed in the patient's data. Such global explanations are mostly given in the form of general and comprehensible rules which can be communicated independent of a specific example \cite{muggleton2018ultra}. With respect to diagnosis it is more often of interest to justify the decision for a particular case in terms of local explanations. A specific type of local explanations are contrastive explanations \cite{dhurandhar2018explanations}. As pointed out by research on how humans explain phenomenons \cite{miller_explanation_2017}, contrastive explanations are an efficient means to highlight which attributes should be present or absent in an example's data to classify it in one way rather than another (``Why class P rather than class Q?''). For instance, it may be highlighted what characterizes pain in contrast to disgust, e.g., the nose is \textit{not} wrinkled and the lids are tightened \cite{schmid2018inductive,schmid_mutual_2020}.

We hypothesize that it might not be enough to assign an example to a class based on the occurrence or absence of attributes. For instance, due to the similarity of facial expressions of pain and disgust, they might be only discernible if temporal relations between different muscle movements are taken into account \cite{wang2013capturing,siebers2016characterizing}, e.g., lowering the brows \textit{overlaps} with closing the eyes in case of an expression of pain in contrast to disgust. In the context of relational domains, it has been pointed out that crucial differences are most suitably demonstrated in the form of near misses \cite{Winston1975}. This involves the selection of appropriate contrastive examples for a concept (near misses) that differ only in small significant characteristics from a target example. Near misses research follows mainly two directions \cite{barnwell2018using}: near miss \textit{generation} \cite{rabold2022generating,hammond2006interactive,gurevich2006active} and near miss \textit{selection} \cite{herchenbach2022explaining}. While methods that generate near misses, usually do this by adapting the target example, near miss selection strategies search the input space in order to find contrastive examples that are most similar to the target example.

In this work we present and compare two approaches that produce contrastive explanations for a pain versus disgust classification task by \textit{selecting} near misses: one approach that takes only \textit{attributes} into consideration and another one that takes \textit{temporal relations} into account. Both approaches have been applied to a real-world data set containing video sequences of facial expressions of pain and disgust. The data set has been obtained from a carefully designed psychological experiment, where facial expressions of pain have been induced through \textit{transcranial magnetic stimulation} \cite{karmann2016role} in human participants in contrast to expressions of disgust. Recording the participants resulted in a data set of pain and disgust video sequences, where each facial expression in a sequence was labelled as an action unit (AU) in accordance to the well-established Facial Action Coding System (FACS) \cite{Ekman1976}. The experiment has been reported in detail in \cite{karmann2016role}.

We apply Inductive Logic Programming (ILP) to obtain an interpretable classification model for each group of video sequences (pain and disgust). For this purpose all video sequences with their annotations have been translated into examples represented as Horn clauses. We train a model for our attribute-based approach (taking into account only the occurrences of AUs) and for our relational approach (where examples are enriched by temporal relations between AUs according to the Allen calculus \cite{Allen1984}). Correctly classified examples are considered for near miss explanations. In both settings, we determine the near misses by ranking contrastive examples according to two different similarity measures. We apply the Jaccard- and the Overlap-coefficient \cite{Jaccard1901,Bradley2006} due to their efficiency in comparing sets. In order to enable set-based similarity computation, we utilize loss-free propositionalisation to the Horn clauses \cite{kuhn2019identifying}.
We hypothesize that near miss explanations yield shorter and more relevant explanations in contrast to less similar examples (far misses) due to the bigger intersection with target examples. We further hypothesize that temporal relations help to distinguish pain from disgust, where attribute-based methods do not suffice. We therefore evaluate the quality of explanations generated by both approaches with respect to contrasting pain with disgust. We compare the length of explanations as an indicator for information aggregation and discuss the separability of pain and disgust by temporal relations.

The paper is organized as follows: in section \ref{sec:related_work} we first give an overview of existing works on near miss explanations. Afterwards we describe both our similarity-based near miss selection approaches, one for attribute-based and one for relational input in section \ref{sec:approach}. In section \ref{sec:evaluation} we present the results of our evaluation. In section \ref{sec:conclusion} we conclude with a discussion and summary of our findings, and outline future prospects for the utilization and extension of our work.

\section{Contrastive Explanations with Near Misses}\label{sec:related_work}

According to Gentner's and Markman's seminal work about structure mapping in analogy and similarity, differences help humans to distinguish between multiple concepts, especially if the concepts are similar \cite{gentner1997structure}. Therefore, contrastive explanations can help making sense of a classifier's decision based on differences between examples that do not belong to the same concept. As motivated in the introduction, contrastive explanations can be generated based on \textit{near misses}. Near misses are defined as being \textit{the examples most similar} to a given example, which do \textit{not} belong to the same class as the given example due to few significant differences \cite{Winston1975}. The differences obtained can be seen as representative characteristics of an example of a class in contrast to the near misses. Contrastive explanations can be derived with the help of near misses in two particular ways  \cite{barnwell2018using}. One approach is, to create an artificial example by making a smallest possible change to the example that is explained (near miss generation), so that it gets classified as an example of another class. The second approach is, to select the most similar example from an opposite class and contrast it with the example that is explained (near miss selection).

Hammond et al. \cite{hammond2006interactive} propose a generative approach for the use case of interface sketching, where interfaces are automatically created based on sketches of shapes, like arrows and boxes, provided by users. To facilitate the process, the authors' approach generates near misses for the given descriptions as suggestions to the users. Although this work is implemented based on Winston's seminal work on near misses \cite{Winston1975}, it heavily depends on pre-defined constraints, which might not always be feasible. Another method that was introduced by Gurevich et al. \cite{gurevich2006active} applies a random sequence of at most 10 different so-called modification operators to positive examples (images of objects belonging to a target concept) in order to generate negative examples, the near misses. This approach can be applied in domains, where modification operators can be easily defined, such as functions over physical properties. Another approach to generating near miss explanations was published lately by Rabold et al. \cite{rabold2022generating}. Their method explains a target example by modifying the rule(s) that apply to it taken from a set of rules learned with Inductive Logic Programming. Each target rule can be adapted by pre-defined rewriting procedures such that it covers a set of near misses, but not the target example. The grounding of the adapted rule serves then as a contrastive explanation.

For some domains it can be tedious or even impossible to pre-define constraints or modification operators and policies. State-of-the-art concept learning often uses feature selection mechanisms to generalize on most relevant features from the feature space, e.g., with the help of statistical methods \cite{KIRA1992249}. However, such techniques may not find the smallest possible subset of relevant features of a concept and are often not designed for explaining classification results. A recently published approach \textit{selects} a set of near miss examples based on similarity metrics, Euclidean, Manhattan and Cosine distance in particular, for the task of image classification and explanation \cite{herchenbach2022explaining}. Even though their method looks promising, it cannot be applied to relational data due to the metric feature space.

In this paper, we present an approach that is based on Winston's definition of a near miss, proposing near miss \textit{selection} for relational data. This way, we want to circumvent the need for constraints and modification operators, since defining them may not be a trivial task in medical use cases, such as pain assessment. To select suitable examples from a contrastive class, we rank the set of contrastive examples based on their similarity to a target example. In order to keep the computational cost of the ranking algorithm low, we propositionalize the relational input and apply two set-based similarity metrics to the examples: the Jaccard- and the Overlap-coefficient. To evaluate whether our proposed approach is useful for explaining pain in contrast to disgust, we apply the similarity-based method for attributes (AUs) and for relations (temporal relations between AUs). Since, according to Winston, contrastive explanations should describe only significant differences, we compare the length of near miss and far miss explanations for both approaches. In the following section we introduce how we produce explanations for interpretable video sequence classification and present our near miss selection approaches.

\section{An Interpretable Approach to Explain Facial Expressions of Pain versus Disgust}\label{sec:approach}

Similar to Rabold et al. we learn a model, a set of rules, for a target class (either pain or disgust) with Inductive Logic Programming (ILP). ILP supports relational explanations in contrast to visual explanations that may not express more than just the presence or absence of features \cite{bach2015pixel,herchenbach2022explaining}. ILP is suitable for inducing rules from data sets that are limited in size \cite{cropper2020turning} and due to its inherent transparency it is applicable to decision-critical domains, e.g., medicine \cite{schmid_mutual_2020}, where comprehensibility of results and accountability play a key role. Explanations generated based on ILP can be tailored to the information need of medical staff, e.g., with verbalized global and local explanations as recently presented in Finzel et al. \cite{finzel2021multimodal}. The following subsections introduce our similarity-based method for ILP to explain facial expressions of pain in contrast to disgust.

\subsection{Learning from Symbolic Representations of Video Sequences}

The input, internal program structure and output of an ILP classifier is represented in expressive first-order logic, in Prolog. The symbolic representation in the form of predicates and implications makes ILP a comprehensible approach, related to logical reasoning processes in humans. The foundations of ILP have already been introduced by Stephen Muggleton in 1991 \cite{muggleton_inductive_1991}. Since then, ILP has been applied to various problems of relational learning \cite{cropper2020turning}.

In ILP clauses must conform to \textit{Horn clauses} and can be written in the form of an implication rule as
\begin{align*}
    A_0 \leftarrow A_{1} \wedge \ldots \wedge A_{n}, \textit{where}\ n \geq 0. 
\end{align*}

\noindent
$A_0$ is called the \emph{head} of the rule and the conjunction of elements
$\bigwedge_{i=1}^n A_i$ is called the \emph{body} of the rule. In Prolog programs that contain predicates and no propositions, each element $A_i$ has the form $p(a_1,\ldots,a_m)$, consisting of a predicate
$p$ that takes arguments $a_i$. An argument is either a constant, a variable or a complex term itself (a predicate with arguments).

Prolog programs consist basically of \textit{facts} (Horn clauses without a body) and may further contain \textit{rules}. While facts state \textit{unconditionally} true relations, e.g.,``a penguin is a bird'', rules state what is \textit{conditionally} true, like ``birds can fly'' (unless they are penguins). In Prolog, implication $\leftarrow$ is denoted by ":-".

Having introduced the basic terminology and syntax, we now define ILP's learning task. ILP learns a set of Horn clauses, also called a \textit{theory}, from a set of positive examples $E{^+}$ that belong to the target concept and a set of negative examples $E{^-}$ that do not belong to the target concept, given background knowledge $B$ about all examples. The goal is to learn a theory $T$ such that, given $B$, $T$ covers all and no negative examples by fulfilling the following conditions:
\begin{itemize}
\item $\forall$ $e$ $\in$ $E{^+}$ : $B$ $\cup$ $T$ $\models$ $e$ ($T$ is \textit{complete})
\item $\forall$ $e$ $\in$ $E{^-}$ : $B$ $\cup$ $T$ $\not\models$ $e$ ($T$ is \textit{consistent})
\end{itemize}

\noindent
To learn a theory based on the symbolic representations of pain and disgust video sequences, we applied the ILP framework \textit{Aleph} \cite{srinivasan_manual}. Its default algorithm performs four steps. First, an example is selected from $E{^+}$. Then, a most specific clause is constructed based on $B$ respectively. In the third step, Aleph searches a clause that is more general than the most specific clause, and finally, all the examples that are covered by the more general clause are removed from the set of remaining positive candidate examples.

\begin{figure}[bt!]
	\centering
	\includegraphics[width=0.86\textwidth]{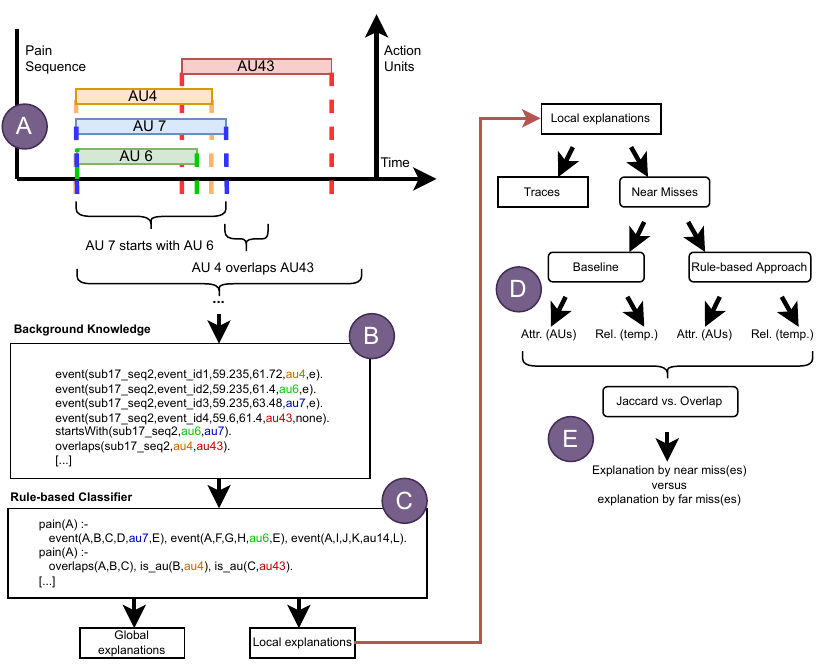}
	\caption{An overview of our approach, showing an illustration of a video sequence of pain (A), an excerpt from the background knowledge in Prolog (B), a subset of the learned rules (C), the implementation (D) and the evaluation setting (E).}
	\label{fig:overview}
\end{figure}

Figure \ref{fig:overview} illustrates the symbolic representations of video sequences in the form of background knowledge and an exemplary learned rule in Prolog from which local and global explanations can be derived. Part A depicts the occurrence of AUs in terms of intervals along the time line of a pain video sequence (AU 4 = \textit{brow lowerer}, AU 6 = \textit{cheek raiser}, AU 7 = \textit{lid tightener}, AU 43 = \textit{eyes closed}, according to \cite{Ekman1976}). Based on the Allen calculus \cite{Allen1984}, which defines an algebra for formal representation and reasoning on time intervals, temporal relations are extracted. For example, AU 7 starts with AU 6 (obviously, the start time of both AUs is the same) and AU 4 overlaps with AU 43. The given sequence is transformed into Prolog predicates, the background knowledge (part B). AUs are represented as events here, where each event has a sequence identifier as a first argument, followed by an event identifier, a start and an end timestamp as well as an indicator for the intensity of the emotion or pain (if applicable for the AU, ranging from a to e). Furthermore temporal relations between AUs are added to the background knowledge and denoted by predicates that contain a sequence identifier and the constants for the AUs respectively.

In a next step (part C), a rule-based model is induced from positive and negative examples and the background knowledge. Our implementation learns rules either on attributes or temporal relations. In the attribute-based approach a rule like the first one, was learned. It consists of event predicates and AUs have been included as constants (letters in lower case). The second rule was learned in the relational setting. It consists of temporal relation predicates and additional predicates to denote AUs. The first rule indicates that persons show pain in a video sequence if AU 7, AU 6 and AU 4 occur. The second rule applies to all pain sequences, where AU 4 overlaps AU 43.

The rule-based classifier can be used to produce global as well as local explanations. A global explanation can be obtained by transforming rules into verbal statements. For local explanations, a rule can be used to query each covered pain example. Therefor variables in a rule (letters in upper case) get grounded with the values from an example's background knowledge. This process is called \textit{tracing}. The computed traces can be considered as being local explanations for the classification of examples \cite{finzel2021multimodal}, yet, they are not contrastive. Ideally, a rule separates examples well and its traces may suffice as explanations. However, where data is very similar, the decision boundary gets thin, meaning that false classifications may occur more often. Contrastive explanations may better describe its borders. The following subsection presents for both our approaches (attribute-based vs. relational), how we produce and evaluate contrastive explanations based on near misses (part D and E, Figure \ref{fig:overview}).

\subsection{Selecting Attribute-based and Relational Near Misses based on Similarity Metrics}

In order to explain what characterizes a pain video sequence in contrast to a disgust video sequence of human facial expressions, we train two ILP models. The first is trained solely on the basis of attributes (AU events). To check, whether including temporal information is beneficial in terms of separating classes, we train another model for relations (temporal relations) between AUs. With the help of the learned models, we produce traces of all examples that we want to contrast. This way, we aim to ensure, that the characteristics, that are highly relevant to separate the different classes, are contained in the explanations.

Both approaches work with finding the most similar example(s) of the opposite class (disgust sequence(s)) first and computing the differences between the target example and the near miss(es) afterwards. The examples have a complex structure and can be represented as graphs. However, computing the similarity between graphs is not trivial. We therefore decided to reduce the problem to computing the similarity between sets. We apply two different similarity metrics: the Jaccard index and the Overlap coefficient (see part E of Figure \ref{fig:overview}).

The Jaccard index \cite{Jaccard1901}, shown in equation \ref{jaccard}, is a similarity measure, where the size of the intersection of two sets, $A$ and $B$, is divided by the size of the union of both sets. The Overlap coefficient divides the size of the intersection of two sets by the size of the smaller set \cite{Bradley2006}. This measure is shown in equation \ref{overlap}.

\begin{equation}
\frac{\left|{A} \cap {B}\right|}{\left|{A} \cup {B}\right|} \label{jaccard}
\end{equation}

\begin{equation}
\frac{\left|{A} \cap {B} \right|}{ min(\left|{A}\right|, \left|{B}\right|)} \label{overlap}
\end{equation}

In order to represent the examples from the data sets as sets, we propositionalize each example's trace. For the attributes, we propositionalize the AU events, such that a feature is composed of an AU, its intensity and an index that indicates the number of occurrence (so that graphs in which an AU occurs multiple times are not reduced to sets that are similar to those of graphs, where the same AU occurs less times). The features for the relations consist of time relation between two AUs and an index that indicates the occurrence as well. % We further add an index to the features to indicate, which of the classifier rules has yielded the trace of the example. - index should be removed, since it does not matter which rule was responsible?

\section{Evaluation}\label{sec:evaluation}

We evaluated which metric yields desirably less near misses for the attributes and relations and whether the explanation length decreases with increasing similarity, by comparing the length of explanations produced by \textit{near} misses versus \textit{far} misses (see Table \ref{tab1}). We compared our results to a baseline that takes the whole background knowledge per example into account for contrasting instead of just taking the traces. We further examine the separability of the trained classifiers.

The attribute-based classifiers trained on the TMS data set \cite{karmann2016role} reached an accuracy of 100 \% for pain, covering 37 positive examples, and an acc. of 100 \% for disgust, covering 93 pos. examples. The relational classifiers reached an acc. of 100 \% for pain, covering 37 pos. examples, and an acc. of more than 60,76 \% for disgust, covering 42 pos. examples.

\begin{table}[]
\centering
\caption{Results for the avg. number of near misses $\bar{n}_{NM}$ per target example and per approach as well as the avg. explanation length for near misses $\bar{L}_{NM}$ and far misses $\bar{L}_{FM}$ in comparison to a baseline for each similarity metric (Jaccard vs. Overlap).} \label{tab1}
\begin{tabular}{p{4.5cm}lllllll}
\cline{1-8}
 & \multicolumn{3}{c}{Jaccard}  &           \space{ } \space{ }               & \multicolumn{3}{c}{Overlaps}                           \\ 
 & \multicolumn{1}{c}{$\bar{n}_{NM}$} \space{ } & \multicolumn{1}{c}{$\bar{L}_{NM}$} \space{ } & $\bar{L}_{FM}$ &  & \multicolumn{1}{c}{$\bar{n}_{NM}$} \space{ } & \multicolumn{1}{c}{$\bar{L}_{NM}$} \space{ } & $\bar{L}_{FM}$ \\ \hline
 
\multicolumn{1}{c}{Attributes (baseline)} & \multicolumn{1}{l}{\textbf{1,09}} & \multicolumn{1}{l}{\textbf{7,92}} & 18,92 &  & \multicolumn{1}{l}{2,36}  & \multicolumn{1}{l}{8,69}  & 12,33 \\ \hline

\multicolumn{1}{c}{Relations (baseline)} & \multicolumn{1}{l}{\textbf{1,19}} & \multicolumn{1}{l}{\textbf{50,68}} & 99,88 &  & \multicolumn{1}{l}{3,06} & \multicolumn{1}{l}{69,76} & 57,49 \\ \hline

\multicolumn{1}{c}{Attributes (our approach)}  & \multicolumn{1}{l}{\textbf{1,75}} & \multicolumn{1}{l}{\textbf{4,52}}  & 6,72  &  & \multicolumn{1}{l}{3,58}  & \multicolumn{1}{l}{4,65}  & 6,21 \\ \hline
\end{tabular}%
\end{table}

The results in Table \ref{tab1} show that near misses (most similar examples) yield considerably shorter contrastive explanations than far misses (least similar examples) for all settings. Although the average number of near misses is lower for the attribute-based baseline compared to our attribute-based approach, the explanation length computed on traces of trained classifiers is much lower for learned attributes compared to whole sequences. For our relational approach, no near misses could be found, since the classifiers learned for pain and disgust had no intersecting features; thus no similarity could be measured. All in all, the Jaccard coefficient produced less near misses and shorter explanations than the Overlap coefficient (highlighted in bold font).

\section{Discussion and Conclusion}\label{sec:conclusion}

We presented two interpretable approaches to distinguish facial expressions of pain and disgust with the help of attribute-based and relational near misses. We computed contrastive explanations based on near miss selection, since, for the given real-world data set, we could not ensure that artificially produced near misses would yield valid real-world examples. To find near misses, we applied the Jaccard and the Overlap coefficient. Our results show that near miss explanations are shorter than far miss explanations and thus yield more efficient contrastive explanations. We found that contrastive explanations are more concise for Jaccard compared to the Overlap coefficient. Furthermore, comparing propositionalized traces gained from the rule-based classifiers, produce much shorter explanations compared to a baseline approach, where examples are considered as a whole, which is not surprising, but an additional validation of our method. For the approach of relational near misses the intersection between examples was empty, which is not a weakness of our approach, but rather indicates the separability of pain and disgust with the help of temporal relations.

One limitation of the presented results is however, that the explanations learned for pain versus disgust may not apply in general. Facial expression data bases are usually small, since the effort of labelling the video sequences is big. Only the data is used for analysis, where multiple FACS experts came to the same label decision. A particular challenge lies in the diversity of the data and in the similarity between emotional states \cite{kunz2013pain}. High quality data bases for facial expressions of pain are rather sparse (e.g.,\cite{lucey2011painful}) and often do not contain further emotions for contrasting. We therefore plan to examine a new data set that is currently curated by psychologists but not published yet.

The two approaches presented in this paper complement existing work on contrastive explanation with near misses by providing an interpretable solution using ILP without the necessity to define contraints or modification operators. We introduced a method for near miss selection based on two computationally simple similarity metrics and applied it to efficiently explain pain versus disgust. We currently plan experiments to evaluate how the explanations help in teaching concepts and how the framework
could be enhanced by further similarity measures and modalities, such as visuals and interaction \cite{bach2015pixel,finzel2021multimodal}. The ultimate goal is, to provide actionable and flexible explainability in the medical domain.

\bibliographystyle{splncs04}
\bibliography{references}
\end{document}